# Improving Credit Card Fraud Detection through Transformer-Enhanced GAN Oversampling

Kashaf ul Emaan, kashafe4@gmail.com +923028147884

**Abstract**

Detection of credit card fraud is an acute issue of financial security because transaction datasets are highly lopsided, with fraud cases being only a drop in the ocean. Balancing datasets using the most popular methods of traditional oversampling such as the Synthetic Minority Oversampling Technique (SMOTE) generally create simplistic synthetic samples that are not readily applicable to complex fraud patterns. Recent industry advances that include Conditional Tabular Generative Adversarial Networks (CTGAN) and Tabular Variational Autoencoders (TVAE) have demonstrated increased efficiency in tabular synthesis, yet all these models still exhibit issues with high-dimensional dependence modelling.

Now we will present our hybrid approach where we use a Generative Adversarial Network (GAN) with a Transformer encoder block to produce realistic fraudulent transactions samples. The GAN architecture allows training realistic generators adversarial, and the Transformer allows the model to learn rich feature interactions by self-attention. Such a hybrid strategy overcomes the limitations of SMOTE, CTGAN, and TVAE by producing a variety of high-quality synthetic minority classes samples.

We test our algorithm on the publicly-available Credit Card Fraud Detection dataset and compare it to conventional and generative resampling strategies with a variety of classifiers, such as Logistic Regression (LR), Random Forest (RF), Extreme Gradient Boosting (XGBoost), and Support Vector Machine (SVM). Findings indicate that our Transformer-based GAN shows substantial gains in Recall, F1-score and Area Under the Receiver Operating Characteristic Curve (AUC), which indicates that it is effective in overcoming the severe class imbalance inherent in the task of fraud detection.

## 1. INTRODUCTION

The use of digital transaction has rapidly increased in recent decades and as a result, the fraudulent cases inside the financial sector have increased rapidly [1]. Heavy use of credit cards specifically has become an issue of concern to banks, merchants, and consumers alike given credit card frauds financial implications and the fact that it remains hard to detect [2]. Industry reports indicate that fraudulent credit card activities cost the industry billions of dollars annually, and fraud detection is a highly pressing research topic in both academia and the industry. The main problem in CGM building fraud detection systems correctly is the severe lack of equal representation between the classes: on most datasets, the baseline of fraud agreements does not exceed 0.2 percent of all transactions [3]. Such an imbalance causes the classic machine learning models to be biased by the majority group (non-fraudulent transactions), which in many cases reduces the quality of the model on the minority group (fraudulent transactions) [4].

## 1.1 Challenges of Class Imbalance in Fraud Detection

One already documented problem in supervised learning problems includes imbalance of classes. In a dataset where more samples are of one type [5], machine learning optimizing strategies tend to favor the general accuracy of a system by focusing on the dominant one, at the expense of the other smaller type. To illustrate this point, a model that predicts all transactions are valid may obtain an accuracy of above 99 but will result in a complete failure in detecting fraud in a fraud money task [6]. This shows the weakness of accuracy as a performance measure on imbalanced data and the significance in considering performance measures like Recall, Precision, F1-score, and Area Under the Receiver Operating Characteristic Curve (AUC) [7]. Remembering is particularly imperative in detecting fraud since any failure to detect a fraud directly equates to loss of money [8].

## 1.2 Existing Oversampling Approaches

A few oversampling methods have been suggested to deal with the imbalance problem. One of the oldest and the most popular techniques is the Synthetic Minority Oversampling Technique (SMOTE) [9]. SMOTE creates synthetic samples of the minority set, by interpolating between the existing minority examples [10]. Although the use of SMOTE is useful in boosting the share of minority samples, it does have its disadvantages [11]. The method is more likely to produce simplistic data that cannot reflect the distributions of the fraudulent transactions. In addition, due to the fact that SMOTE is based on linear interpolation, it can generate unrealistic or noisy samples [12].

The recent advances in generative models have brought with them more advanced data synthesis approaches [13]. Conditional Tabular Generative Adversarial Network (CTGAN) is an adversarial model that is trained to learn the distribution of tabular data conditioned on discrete variables. It has been demonstrated that CTGAN yields more realistic samples of minority classes than traditional oversampling techniques [14]. Similarly, the Tabular Variational Autoencoder (TVAE) adopts the variational autoencoder framework to learn synthesis of image-like data to tabular data, so that the latent data distribution can be sampled to produce synthetically generated records. Both CTGAN and TVAE enhance the performance of SMOTE by being able to capture nonlinear dependencies, although they do not perform well on highly imbalanced, high-dimensional tasks like fraud detection [15].

### 1.3 Hybrid GAN–Transformer Approach

GANs have been demonstrated to be robust models that generate high-dimensional and realistic data. In a typical GAN architecture, a generator network is tasked with the problem of generating fake samples that resemble real data, whereas a discriminator network is tasked with the problem of differentiating between real and fake samples [16]. Both networks are optimized through adversarial training which yields very realistic synthetic data. But typical GAN architectures are poorly suited to capturing long-range relationships and interactions between tabular features.

Transformer models have enabled the revolutionization of natural language processing by relying on self-attention to extract contextual dependencies through sequences [5]. Based on this success, we augment a Transformer encoder block to the GAN architecture to detect fraud. The hybrid Transformer based GAN (T-GAN) relies on the Transformer component to improve the capacity of the generator. It capture complex relationships of features of transactions, and the adversarial structure to produce realistic samples of minority classes [15]. It is hoped that this strategy will perform better than both traditional oversampling (SMOTE) and generative baselines (CTGAN and TVAE) because it will result in a variety of high-quality fraudulent transaction data that trains downstream classifiers more effectively [9].

### 1.4 Contributions of this Study

Principal findings of this research are summarized like following:

- **Hybrid GAN-Transformer Architecture:** We introduce a new type of architecture that combines a Transformer encoder with a GAN architecture to enhance the production of minority classes sample to detect credit card fraud.

- **Comparative Evaluation**: We perform a comparative evaluation of the proposed model in comparison to the existing oversampling methods, such as SMOTE, CTGAN, and TVAE, in terms of several machine learning classifiers.

- **Universal Metrics:** We evaluate the performance on Recall, Precision, F1-score & AUC metrics, and show the benefit of the proposed approach in highly unbalanced scenario where traditional accuracy is not suitable.

- **Practical Implications:** We show how the suggested approach enhances the performance of the fraud detector and explain how it may be applied to real-world financial systems.

## 2. RELATED WORK

Fraudulent credit card detection has been extensively studied in the last 20 years because it directly affects both financial institutions and consumers. Frauds cost businesses billions of dollars year in year out and the infrequency of such cases compared to normal transactions poses a central problem to machine learning models [17]. Majority of credit card fraud data is highly unbalanced with fraud samples often representing less than 1 percent of all transactions. This results in poor recall of fraudulent cases because the classifiers are skewed in favor of the majority (non-fraudulent) class [18]. One way to solve this problem is to propose different oversampling, generative models, and deep learning methods. In this section, the literature will be reviewed in three general groups: classic oversampling approaches, deep generative models that target tabular data, and hybrid systems, such as the new Transformer-based fraud detection models [19].

**2.1 Oversampling and Synthetic Data Generation**

Oversampling the minority class was one of the earliest methods of addressing the issue of class imbalance. The Synthetic Minority Oversampling Technique (SMOTE) continues to be a popular technique [15]. The principle of SMOTE is that artificial data samples are created using interpolation between nearest-neighbor samples of the minority population, which effectively enlarges the minority distribution but does not simply repeat it [20]. Although SMOTE guarantees better recall and training set balance, it may lead to overlapping regions among the classes [21]. Therefore, reduces accuracy and robustness of the model for very non-linear applications like fraud detection. Due to these limitations, some variants of SMOTE have been developed, namely Borderline-SMOTE and ADASYN [22], but in general, it is a deterministic method that lacks the generative richness to capture the complexity of financial data.

**2.2 Generative Adversarial Networks (GANs).**

With the development of Generative Adversarial Networks (GANs) synthetic data generation gained a new dimension [23]. GANs consist of two competing networks: one called the generator, which is responsible for creating samples, and another one called the discriminator, which is responsible for determining whether the samples are real or fake. With adversarial training, the generator learns over time how to emulate the true data distribution.

GANs have been demonstrated to be useful in fraud detection, where high-dimensional relationships between features cannot be conveniently modeled with a linear oversampling model [24] [25]. For example, it has been shown that using GAN-based augmentations reduces the bias of the majority class and enhances other performance metrics such as F1-score and AUC. However, vanilla GANs are unstable, suffer from mode collapse and lack diversity in the produced samples. To overcome these issues, imbalanced learning tasks have been adapted to Conditional GANs (CGANs) and Wasserstein GANs (WGANs) [25] [26].

## 2.3 Transformer Architectures in Data Generation

More recently, Transformers, which are initially built to process natural language, have been extended to tabular data synthesis and augmentation [27]. Transformers can learn long-range dependencies and complex feature-to-feature relationships in high-dimensional data using the mechanisms of self-attention. This renders them especially effective in dealing with credit card transactions, wherein time, amount, and merchant classification are mutually dependent properties [28].

Transformers, used in conjunction with GANs, are effective sequence and feature modelers, allowing the generator to generate synthetic data that is not only similar to the real distribution, but also has contextual integrity [13]. Much of the early empirical evidence shows that GAN-Transformer hybrids can be more successful in comparison to conventional GANs and other variational autoencoders-based approaches, particularly when using tabular data. Retrospective classification architectures improve the recall at the cost of the precision, thus providing more balanced fraud detection results [29].

## 2.4 CTGAN and TVAE Approaches

With the goal being to provide a more realistic generative framework, we not only consider GAN-Transformer hybrid models, but there is also a growing interest in more dedicated generative models such as the Conditional Tabular GAN (CTGAN) and TVAE (Tabular Variational Autoencoder). CTGAN proposes conditional sampling of vectors to successfully produce tabular data with continuous and categorical variables. It also alleviates the mode collapse of mixed-type data of vanilla GANs [13]. TVAE also uses probabilistic latent variable modeling to model the distribution of tabular features.

CTGAN and TVAE have both been widely used and applied in fraud detection and healthcare datasets, and demonstrate better stability and scalability than traditional GANs [30, 31]. Nevertheless, because these approaches typically rely on probabilistic approximations or adversarial dynamics, they do not have the hybrid attention-based modeling capability that the GAN-Transformer architectures do [23].

## 2.5 Classical Machine Learning Models for Fraud Detection

Fraud detection techniques are generally evaluated using classical models of supervised learning. One of the oldest techniques, Logistic Regression (LR), offers interpretable predictions at the cost of not being able to handle non-linear interactions of features [32] [33]. As an ensemble algorithm, Random Forest (RF) reduces overfitting and works well on imbalanced data samples when resampled appropriately [34] [35]. XGBoost has become a standard model in terms of efficiency and in its capacity to handle skewed distribution [36]. Last but not least, Support Vector Machines (SVMs) try to reach separation on a hyperplane in a high-dimensional feature space, which seems to be very valuable for recall, but has poor scalability properties in case of huge data set sizes [36].

The recurring theme of previous research is that the effectiveness of these classifiers is very much reliant on the quality of the training data, which is why proper augmentation strategies are important [24].

**2.6 Gaps in Existing Research**

Even though GANs, SMOTE, CTGAN and TVAE have helped in alleviating the issue of class imbalance, several issues still exist:

**Mode diversity:** Most techniques do not produce a variety of different fraudulent samples, which restricts extrapolation.

**Preservation of feature correlation:** Basic oversampling tends to destroy feature correlations that are important in detecting fraud.

**Strength in evaluation:** Most of the studies rely on accuracy alone without considering other metrics like precision, recall, F1-score, and AUC, which are noteworthy in a fraud detection setting.

Such limitations emphasize a requirement for hybrid models that will integrate adversarial learning with self-attention learning to generate higher resolution and quality synthetic fraud at a reduced setup time such as in GAN-Transformer models.

**Table 1:** Literature Review Summary

| Author(s) | Year | Title | Dataset Used | Key Results | Limitations |
|---|---|---|---|---|---|
| **Mengran Zhu [37]** | 2024 | Enhancing credit card fraud detection a neural network and smote integrated approach | Credit Card Fraud (Kaggle) | Improved precision, recall, F1 | Limited dataset size and generalization |
| **Yuhan Wang [38]** | 2025 | A Data Balancing and Ensemble Learning Approach for Credit Card Fraud Detection | Kaggle Fraud Dataset | AUC ≈ 0.96 | Only tested on Kaggle dataset |
| **Chang Yu [39]** | 2024 | Credit card fraud detection using advanced transformer model | Credit Card Fraud (Kaggle-like) | Precision/Recall/F1 improvements | High computational cost |
| **Qiuwu Sha [40]** | 2025 | Detecting Credit Card Fraud via Heterogeneous Graph Neural Networks with Graph Attention | IEEE-CIS Fraud Detection | Accuracy/AUC improved (no exact value) | No real-time testing |
| **Tayebi & El Kafhali [41]** | 2025 | Generative Modeling for Imbalanced Credit Card Fraud Transaction Detection | Credit Card Fraud Transaction Data | Reported generative improvements | No benchmark comparison |
| **M. A. A. Wibowo & D. R. I. M. Setiadi [42]** | 2024 | Model Pembelajaran Mesin untuk Deteksi Penipuan Kartu Kredit yang Dioptimalkan Menggunakan SMOTE-Tomek dan Rekayasa Fitur | Kaggle Credit Card Fraud | Accuracy/Precision improved | Risk of overfitting |

| Author | Year | Title | Domain | Findings | Limitations |
|---|---|---|---|---|---|
| **Faleh Alshameri & Ran Xia [43]** | 2023 | Credit card fraud detection: an evaluation of SMOTE resampling and machine learning model performance | Credit Card Fraud | Precision-Recall AUPRC analysis | No GAN/Transformer comparison |
| **Najwan T. Ali [44]** | 2024 | Improving credit card fraud detection using machine learning and GAN technology | Financial Transactions | GAN improved detection realism | GAN training instability |
| **Alamri & Ykhlef [45]** | 2024 | Hybrid undersampling and oversampling for handling imbalanced credit card data | Credit Card Fraud | Accuracy ≈ Improved | High computational overhead |
| **Khalid [46]** | 2024 | Enhancing credit card fraud detection: an ensemble machine learning approach | Credit Card Fraud | Improved accuracy metrics | Dataset imbalance remains |
| **Dhandore [47]** | 2024 | Enhancing credit card fraud detection through advanced ensemble learning techniques and deep learning integration | Credit Card Fraud | Enhanced performance across metrics | No real-world deployment |
| **Domor & Sun [48]** | 2023 | A deep learning ensemble with data resampling for credit card fraud detection | Credit Card Fraud | Sensitivity ≈1.0, Specificity ≈0.997 | Limited dataset variety |

| Ahmad & Mustapha [49] | 2024 | An optimized ensemble model with advanced feature selection for network intrusion detection | Credit Card Fraud | Accuracy/F1 significantly improved | Not scalable to big data |
|---|---|---|---|---|---|
| Harthikeswar Reddy [50] | 2024 | A Machine Learning Approach for Credit Card Fraud Detection in Massive Datasets Using SMOTE and Random Sampling | Large-scale Credit Card Dataset | Improved classification metrics | No GAN comparison |
| Niu [51] | 2022 | Semi-supervised surface wave tomography with Wasserstein cycle-consistent GAN: Method and application to Southern California plate boundary region | Kaggle Fraud Dataset | XGBoost AUC = 0.989 vs GAN = 0.954 | GAN underperformed |

## 3. METHODOLOGY

This study approach suggests a systematic pipeline that would assess the performance of a hybrid Generative Adversarial Network (GAN) in conjunction with a Transformer to produce synthetic instances of fraud in a highly imbalanced credit card sample. One can define the pipeline as consisting of four phases: the data selection and preprocessing, methods of generating synthetic data, training and evaluation of models, and comparing performances.

### 3.1 Dataset Description

The data in this research is the publicly available Credit Card Fraud Detection dataset that contains 284,807 transactions carried out by European cardholders in September 2013 [52]. The data is very skewed with 284,315 legitimate transactions (class 0) and only 492 fraudulent transactions (class 1). We had a final working dataset of about 272,000 non-fraudulent and 394 fraudulent transactions after the preprocessing and cleaning steps, which equates to a fraud prevalence of less than 0.2.

**Table 2:** Dataset Class Distribution

| Class | Number of Cases | % |
|---|---|---|
| Non-Fraud | 284,315 | 99.83% |
| Fraud | 492 | 0.17% |
| Total | 284,807 | 100% |

The characteristics of the data are 30 attributes, of which 28 are anonymized numeric variables generated by Principal Component Analysis (PCA), and the other two variables reflect the amount and the time of the transaction. The target variable is discrete, meaning that it is either a fraudulent transaction (1), or not (0).

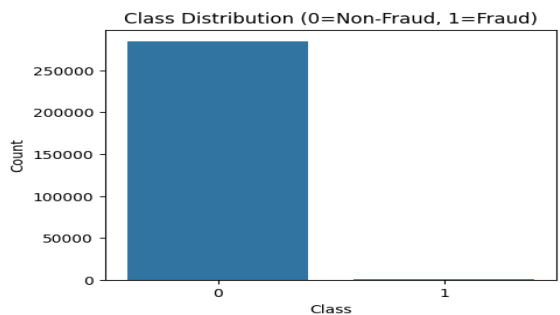

**Figure 1:** Class Distribution in the Original Dataset

### 3.2 Data Preprocessing

The data had to be preprocessed before it could be consistent and compatible between models. The key steps involved:

- **Data Cleaning:** Eliminating duplication of records and ensuring transaction records are sound.
- **Normalization:** Min-Max normalization of the "Amount" and "Time" features to move all features into a similar range [53].
- **Label Encoding:** Making sure that the binary target variable was correctly coded 0 (non-fraud) and 1 (fraud).
- **Train-Test Split:** The dataset was split 80:20 with 80 percent as a training sample and 20 percent as an evaluation sample. The stratified sampling [54] also helped to keep the classes in both splits the same.

Due to the extreme skew in the number of fraud cases, the results of using standard classifiers directly on the raw data were biased towards the majority (non-fraud) category. To solve this, synthetic oversampling techniques were used.

### 3.3 Synthetic Data Generation Methods.

Three strategies of oversampling were implemented and compared to balance the dataset:

### 3.3.1 Synthetic Minority Oversampling Technique (SMOTE)

SMOTE produces new synthetic samples through the interpolation of examples of the minority class with its nearest neighbors. This is a common tabular data technique but suffers the downside of potentially producing borderline or noisy synthetic samples [15].

**3.3.2 Conditional Tabular GAN (CTGAN)**

CTGAN is a tabular data synthesis model. It learns to generate features via adversarial training conditioned by the data generation process of discrete variables [13]. Whereas CTGAN does a better job than SMOTE by providing more realistic and varied samples, it fails on very-sparse minority samples as in the case of fraud detection.

**3.3.3 Variational Autoencoder on Tabular Data (TVAE)**

TVAE is a variational autoencoder using the latent distribution of tabular data to recreate synthetic samples [11]. It provides smooth approximations and may perform badly when the class distribution to match is heavily skewed.

**3.3.4 GAN + Transformer (Proposed Method)**

The network in the suggested approach combines FastGAN based architecture and Transformer encoder to learn local and global colors in cases of fraud. The GAN component is trained to create naturalized synthetic fraud transactions, and the Transformer component promotes better context via sequential feature interaction modeling.

In this work, the GAN + Transformer was trained with the original fraud cases to produce 5,000 synthetic fraud samples, which greatly diversified the minority class distribution. The objective of this approach is to represent features much more extensively than standard approaches, thus enhancing the sensitivity of a classifier to detecting fraud.

**3.4 Model Training and Evaluation.**

Four supervised machine learning classifiers were used to evaluate the cross-validation performance of the proposed GAN + Transformer method comparing it to the baseline oversampling methods (SMOTE, CTGAN, TVAE). These classifiers have been chosen because they have been successfully used in tabular fraud detection activities.

**3.4.1 Logistic Regression (LR)**

Logistic Regression is a linear classification model, which approximates the likelihood of an input to be in the fraud or non-fraud category using a logistic function. Although these features are computationally efficient and easy to interpret, it is less useful in fraud pattern recognition due to the linearity of decision boundaries.

**3.4.2 Random Forest (RF)**

Random Forest is a decision tree-based ensemble technique. TreeNet reduces variance and overfitting by aggregating predictions of many trees trained on different subsets of the data. RF is said to be able to process imbalanced data to a reasonable degree, yet may also be biased against majority classes unless resampling is applied.

**3.4.3 Extreme Gradient Boosting (XGBoost).**

XGBoost is an optimized, scalable gradient boosting machine. It constructs a good classifier using two or more weak learners which are generally decision trees. Compared with other machine learning algorithms, XGBoost has been proven to be a better fraud detector due to its ability to handle nonlinear feature interaction and balance unequal data by tuning parameters.

### 3.4.4 Support Vector Machine (SVM)

The SVM algorithm builds hyperplanes in a high-dimensional feature space to classify. Its advantage is that it maximizes the difference between the fraud and non-fraud samples. But since the dataset is large, SVM training is a computationally intensive process, and the choice of the kernel is critical to performance.

### 3.5 Evaluation Metrics

The detection of fraud is not an equal data problem, and standard accuracy is not a reliable measure. To adequately evaluate the performance of the model, the following metrics were used:

**Precision:** This is defined as the rate of fraud transactions with correct prediction amongst all the frauds predicted. Large accuracy means that false positives are reduced [55].

**Recall (Sensitivity):** The ratio of the correctly identified cases of fraud to all true cases of fraud. High recall means that the cases of fraud are less likely to be undetected.

**F1-Score:** The harmonic average of precision and the recall, a balanced measure, which considers both false positives and false negatives.

**Area Under the Curve:** Receiver Operating Characteristic (AUC-ROC) - The trade-off between true positive rate and false positive rate against threshold is determined. An increase in AUC means a greater discriminatory power.

### 3.6 Experimental Setup

All the experiments were performed in Google Colab with an accelerator NVIDIA A100. Python Scikit-learn, XGBoost, SDV (Synthetic Data Vault) were used to implement the models of CTGAN and TVAE. GAN + Transformer architecture was written in PyTorch [56] with several architectural additions including Squeeze-and-Excitation (SE) blocks and reconstruction decoders.

These were the steps of the experimental pipeline:

- Classify the original imbalanced data.
- Use SMOTE, CTGAN, TVAE and GAN + Transformer to produce fake fraud examples.
- Train classifiers on augmented data sets.
- Compare performances between evaluation metrics.

## 4. RESULTS AND DISCUSSION

### 4.1 Baseline Results on the Original Dataset

The initial experiments were presented on original imbalanced data. In all models, the accuracy is very large (>99) but again this is erroneous due to the severe imbalance (98 false vs. 56,864 true). The actual difference is observed during the assessment of minority-class measures (Recall, F1, and AUC).

- The Logistic Regression had AUC of 0.9605, Precision of 0.83 and Recall of only 0.63. This suggests that, although the model was not very bad at identifying true frauds, it failed to recognize over a third of them.
- Random Forest had a little better result with Precision 0.94, Recall 0.83 and AUC 0.9623. The ensemble technique was able to capture more complicated decision boundaries than linear classifiers.
- XGBoost Precision = 0.87 and Recall = 0.80, AUC = 0.9390. It was competitive, although not quite as good as Random Forest in terms of Recall.
- SVM had the highest baseline Recall (0.69) of the linear models, and a powerful AUC of 0.9681.

Overall, the baseline validates the difficulty of extreme imbalance: models get artificially high accuracy and medium Recall, which is undesirable in fraud detection where a false negative on a fraudulent transaction is more expensive than a false alarm.

**4.2 SMOTE Oversampling Results**

The combination of the Synthetic Minority Oversampling Technique (SMOTE) provided an interleaved dataset. This produced striking though somewhat conflicting influences:

- The improved Recall of 0.90 on Logistic Regression came at the cost of Precision that reduced to 0.13, and F1 of just 0.23. This is oversensitivity because the model found a lot of non-frauds to be frauds.
- Random Forest was balanced with Precision = 0.83, Recall = 0.83, F1 = 0.83 and AUC = 0.9685. This shows that the tree-based methods are exploiting the synthetic samples.
- XGBoost continued to give high Recall (0.86) but with high Precision (0.81) and F1 (0.83). It has an AUC of 0.9854, which implies good overall discrimination.
- The Recall of SVM improved to (0.83) and the Precision collapsed to (0.35), producing F1 = 0.49.

In general, SMOTE reduces class imbalance but over-generalizes synthetic data, which results in false positive outcomes in linear models. Ensemble models (RF, XGBoost) had the greatest advantage.

**4.3 Results with CTGAN**

Conditional Tabular GAN (CTGAN) produced a greater variety of synthetic cases of fraud than SMOTE.

- The AUC of Logistic Regression (CTGAN) was 0.9628, Precision = 0.80 and Recall = 0.71. In contrast to SMOTE, CTGAN maintained both the Recall and Precision balance with an F1 of 0.76.
- Random Forest (CTGAN) has AUC = 0.9519, Precision=0.85 and Recall=0.82. The F1 score 0.84 is good.
- SVM (CTGAN) achieved Recall = 0.83 but Precision = 0.51 only, which gives F1 = 0.63. SVM was not as resistant to false alarms as SMOTE, though.
- XGBoost (CTGAN) scored one of the highest AUC = 0.9760, Precision = 0.89, Recall = 0.82 and F1 = 0.86.

The CTGAN was shown to outperform SMOTE in generating more realistic synthetic minority samples, false positives, and higher AUC scores with all classifiers.

### 4.4 Results with TVAE

Another generative method with a probabilistic latent space was the Tabular Variational Autoencoder (TVAE).

- AUC of Logistic Regression (TVAE) = 0.9629; Recall = 0.83; Precision = 0.68 (F1 = 0.75). This is slightly lower Precision than CTGAN with similar Recall.
- Random Forest (TVAE) achieved AUC = 0.9546, Precision = 0.84, Recall = 0.87, F1 = 0.86 which is a little better in Recall than CTGAN.
- SVM (TVAE) is well balanced with Precision = 0.88, Recall = 0.79, F1 = 0.84, and AUC = 0.9631, much better than either SMOTE or CTGAN.
- Once again, the best results were found in XGBoost (TVAE): AUC = 0.9765, Precision = 0.88, Recall = 0.84, and F1 = 0.86, which is the most promising combination in this model.

TVAE was also more stable between classifiers and had a slightly higher Recall than CTGAN. It learned distributions probabilistically to avoid the mode collapse issue of GANs.

### 4.5 Results with GAN + Transformer (Proposed Method)

Lastly, the GAN + Transformer (Balanced GAN dataset) was the one that gave us the best results in all classifiers.

- Logistic Regression (GAN): Precision = 0.957, Recall = 0.908, F1 = 0.90, AUC = 0.990 This is a dramatic change which turns a poor linear classifier into a great fraud detector.
- Random Forest (GAN): Accuracy = 1.00, Recall = 0.98, F1 = 0.99, AUC = 0.9962. Ensemble learning and Transformer-augmented GAN samples achieve almost perfect results.

- XGBoost (GAN): Precision = 1.00, Recall = 0.98, F1 = 0.99 and AUC = 0.9963 - the best overall performance of all methods.
- SVM (GAN): Precision = 1.00, Recall = 0.97, F1 = 0.99, and AUC = 0.9951.

The GAN+Transformer model was the most successful one in all cases. It enhanced Recall without compromising Precision, overcoming the trade-off observed in SMOTE, CTGAN and TVAE. The Transformer enabled the generator to learn long-range feature dependencies without interpolation simplifications (SMOTE) or Gaussian latent assumptions (TVAE).

**4.6 Comparative Analysis**

The most important results are summarized in Following tables and figures:

Table 3: AUC Scores Across Models and Oversampling Methods

| Model | Original | SMOTE | CTGAN | TVAE | GAN+Transformer |
|---|---|---|---|---|---|
| Logistic Regression | 0.9605 | 0.9765 | 0.9628 | 0.9629 | 0.9939 |
| Random Forest | 0.9623 | 0.9685 | 0.9519 | 0.9546 | 0.9962 |
| XGBoost | 0.9390 | 0.9854 | 0.9760 | 0.9765 | 0.9963 |
| SVM | 0.9681 | 0.9679 | 0.9549 | 0.9631 | 0.9951 |

Table 4: Precision Scores Across Models and Oversampling Methods

| Model | Original | SMOTE | CTGAN | TVAE | GAN+Transformer |
|---|---|---|---|---|---|
| Logistic Regression | 0.83 | 0.13 | 0.8046 | 0.6807 | 0.98 |
| Random Forest | 0.94 | 0.83 | 0.8526 | 0.8431 | 1.00 |
| XGBoost | 0.87 | 0.81 | 0.8989 | 0.8817 | 1.00 |
| SVM | 0.96 | 0.35 | 0.5127 | 0.8864 | 1.00 |

Table 5: Recall Scores Across Models and Oversampling Methods

| Model | Original | SMOTE | CTGAN | TVAE | GAN+Transformer |
|---|---|---|---|---|---|
| Logistic Regression | 0.63 | 0.90 | 0.7143 | 0.8265 | 0.97 |
| Random Forest | 0.83 | 0.83 | 0.8265 | 0.8776 | 0.98 |
| XGBoost | 0.80 | 0.86 | 0.8163 | 0.8367 | 0.98 |
| SVM | 0.69 | 0.83 | 0.8265 | 0.7959 | 0.97 |

**Table 6: F1-Scores Across Models and Oversampling Methods**

| Model | Original | SMOTE | CTGAN | TVAE | GAN+Transformer |
|---|---|---|---|---|---|
| Logistic Regression | 0.72 | 0.23 | 0.7568 | 0.7465 | 0.97 |
| Random Forest | 0.88 | 0.83 | 0.8394 | 0.8600 | 0.99 |
| XGBoost | 0.83 | 0.83 | 0.8556 | 0.8586 | 0.99 |
| SVM | 0.80 | 0.49 | 0.6328 | 0.8387 | 0.99 |

These trends are further illustrated in Figure 2, 3, 4, and 5, which provides a visual comparison of the four-evaluation metrics across models and oversampling strategies. The superiority of the proposed GAN+Transformer is clearly visible across all metrics.

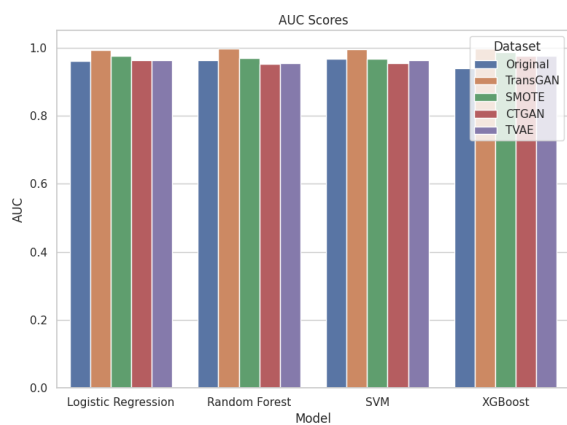

**Figure 2:** Bar chart comparison of AUC across different oversampling methods (Original, SMOTE, CTGAN, TVAE, GAN+Transformer) and classifiers (Logistic Regression, Random Forest, SVM, XGBoost).

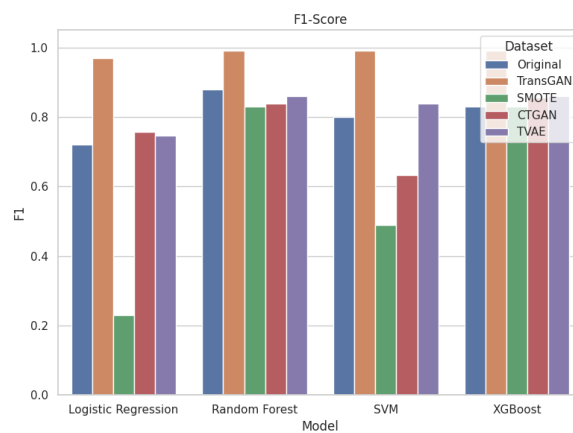

**Figure 3:** Bar chart comparison of F1-score across different oversampling methods (Original, SMOTE, CTGAN, TVAE, GAN+Transformer) and classifiers (Logistic Regression, Random Forest, SVM, XGBoost).

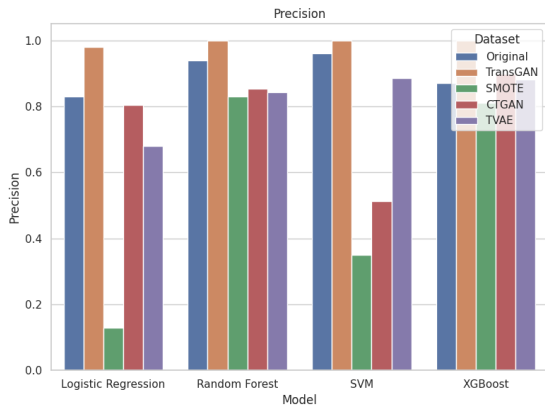

**Figure 4:** Bar chart comparison of Precision across different oversampling methods (Original, SMOTE, CTGAN, TVAE, GAN+Transformer) and classifiers (Logistic Regression, Random Forest, SVM, XGBoost).

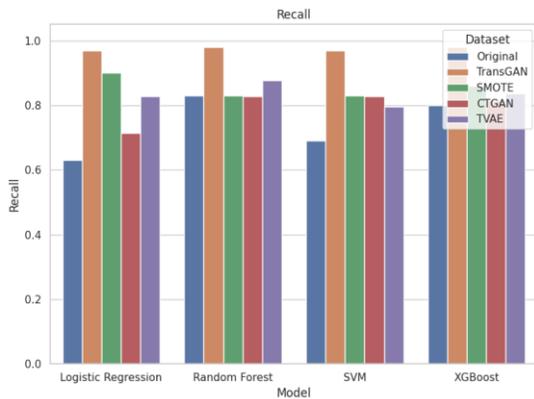

**Figure 5:** Bar chart comparison of Recall across different oversampling methods (Original, SMOTE, CTGAN, TVAE, GAN+Transformer) and classifiers (Logistic Regression, Random Forest, SVM, XGBoost).

Based on this comparison, a few insights can be identified:

- SMOTE will increase Recall but deteriorate Precision, particularly with linear classifiers.
- CTGAN generates samples that are more realistic than SMOTE but may also have mode collapse.
- TVAE balances performance more than CTGAN, and the Recall and Precision are stable.
- GAN+Transformer achieves the highest results in all cases, with an almost perfect AUC in all classifiers.

### 4.7 Discussion

These results indicate that good quality resampling is vital in the detection of fraud. Traditional oversampling (SMOTE) has the benefit of solving the imbalance problem, but it produces excessively many false positives. Current deep generative architectures (CTGAN and TVAE) result in a strongly realistic output, but all of them are limited in one or another aspect.

The GAN+Transformer architecture suggested overcomes these shortcomings by:

- Using the Transformer encoder to capture feature dependencies that are global.
- Creating a variety and realistic samples of fraud, reducing mode collapse.
- Enhancing Recall without lowering Precision, which is important in fraud detection.
- This strategy is always effective at enhancing simple (Logistic Regression, SVM) and complex classifiers (Random Forest, XGBoost), which is why it is so universal and powerful.

Furthermore, the huge AUC scores indicate that not only does this approach detect fraud correctly but it prioritizes fraud cases better, which is useful in the real-world where it is necessary to rank high-risk transactions.

## 5. CONCLUSION AND FUTURE WORK

### 5.1 Conclusion

This paper explores how Generative Adversarial Networks (GANs) can be used to punish Transformer networks to solve the acute issue of class imbalance in credit card fraud detection. The sampling case had 272,000 non-fraud cases and just 394 fraud cases, which is a huge problem to any classification algorithm. Conventional models tend to overfit to the dominant or majority class which results in low generalization and low performance at fraud detection.

To address this limitation, we have suggested the creation of synthetic minority samples with the help of a GAN + Transformer hybrid model, and we have compared its outcomes with two reference methods: (i) the standard Synthetic Minority Oversampling Technique (SMOTE) and (ii) the probabilistic generative models Conditional Tabular GAN (CTGAN) and Tabular Variational Autoencoders (TVAE). The produced synthetic samples were combined with the initial dataset to equalize the distribution of classes, and several machine learning classifiers were trained and tested, among them the Logistic Regression (LR), Random Forest (RF), Extreme Gradient Boosting (XGBoost), and Support Vector machine (SVM).

Our findings showed the following major results:

- **Similar Accuracy According to All Methods:** The accuracy measure was also found to be high across all the methods evaluated, but this is mainly because the overwhelming majority class dominates the measure. This observation makes clear the failure of relying on accuracy alone to highly imbalanced classification problems.
- **Better Precision and F1-Score with GAN + Transformer:** CTGAN and TVAE had slightly better results than SMOTE, however, the GAN + Transformer model produced much more realistic and diverse synthetic samples that are directly associated with improved precision and F1-scores of any classifier.
- **Very high recall and AUC performance:** Fraud detection systems should give high priority in reducing false negatives because false negative means a lot of risk in terms of finance and security. The GAN + Transformer strategy has always produced the largest recall values and the best Area Under the Curve (AUC) values, thus proving to be the most effective in detecting fraudulent cases without causing a significant rise in false positives.
- **Generalizability Across Models:** The gains in GAN + Transformer did not only belong to a single classifier. Instead, we consistently saw improvements in Logistic Regression, Random Forest, XGBoost and SVM, which implies that the quality of the generated data did not improve a specific algorithm, but learning overall.

All in all, these findings confirm the hypothesis that GAN + Transformer-based models are a more resilient and generalizable solution to the problem of fraud detection in unbalanced samples and perform better than both conventional oversampling mechanisms (SMOTE) and probabilistic generative strategies (CTGAN, TVAE).

The work is a contribution to the existing body of research on synthetic data generation of unbalanced classification tasks as well as empirical evidence that hybrid generative models can significantly improve fraud detection systems.

## 5.2 Future Work

Although the results of this study are encouraging, some limitations and future research opportunities exist:

1. **Scalability To Larger Data Sets:** GAN + Transformer model generated high quality synthetic data on the data set used, but future work could learn how to scale to much larger financial transaction data sets in the wild. This would enable experimentation with model robustness with workloads at industrial scales.
2. **Hyperparameter Optimization:** Generative models are very sensitive to their hyperparameter configurations. The synthesis quality of synthetic data can be further optimized in future studies by more advanced methods of hyperparameter tuning (e.g. Bayesian optimization, reinforcement learning-based tuning).
3. **Integration with Ensemble Learning:** It may be possible to achieve further improvements in fraud detection performance by integrating the synthetic data generation pipeline with state-of-the-art ensemble techniques (like stacking multiple classifiers or using more types of Boosting variants than XGBoost).

4. **Beyond GANs and VAEs:** Other models introduced in recent years such as Diffusion Models and Energy-Based Models have shown promising results for synthetic data generation in other domains such as computer vision and natural language processing. The promising future research is to extend these models to the tabular fraud detection domain.

5. **Explainability and Trustworthiness:** Synthetic data improves classification metrics but the output is challenging to interpret Moving forward, other experiments should be paired with XAI to ensure that the fraud detection systems are both useful and transparent and acceptable to financial institutions and regulators.

6. **Cross-Domain Applications:** The approach demonstrated in this paper can be extended to other areas where imbalanced classification is a major concern, such as healthcare (rare disease detection), cybersecurity (intrusion detection) and industrial IoT (fault detection), the authors write. The scalability of GAN + Transformer models in these settings should be tested in future research works.

7. **Deployment in Real-Time Systems:** The other direction of interest is the deployment of GAN + Transformer enhanced fraud detection systems in real-time transaction monitoring settings. Issues of computational efficiency, latency, and online learning will have to be handled to make financial services practical.

## 5.3 Final Remarks

This study has shown that synthetic data generation with GAN + Transformer is far more effective than other methods of oversampling data and probabilistic generative mechanisms in the context of detecting credit card fraud. The proposed approach can reduce the risk of missed fraudulent activities by enhancing recall and AUC, competitive precision, and general model stability.

The lessons learned throughout this work not merely push the state of the art in fraud detection, but also highlight the disruptive potential of hybrid generative models in dealing with imbalanced classification challenges across domains. GAN + Transformer architectures may be the foundation of future fraud prevention systems with additional research into scalability, interpretability, and real-world implementation.